# Learning Transferable Policies for Monocular Reactive MAV Control

Shreyansh Daftry, J. Andrew Bagnell, and Martial Hebert

Robotics Institute, Carnegie Mellon University, Pittsburgh, USA
{daftry,dbagnell,hebert}@ri.cmu.edu

**Abstract.** The ability to transfer knowledge gained in previous tasks into new contexts is one of the most important mechanisms of human learning. Despite this, adapting autonomous behavior to be reused in partially similar settings is still an open problem in current robotics research. In this paper, we take a small step in this direction and propose a generic framework for learning transferable motion policies. Our goal is to solve a learning problem in a target domain by utilizing the training data in a different but related source domain. We present this in the context of an autonomous MAV flight using monocular reactive control, and demonstrate the efficacy of our proposed approach through extensive real-world flight experiments in outdoor cluttered environments.

## 1 Introduction

Micro Aerial Vehicles (MAVs) are becoming increasingly popular in number of important applications [3]. As these robots aspire for long-term autonomous operation in unstructured environments, designing hand-engineered perception and control software remains a tedious process, even for basic tasks like collision avoidance [5,2]. However, in recent years learning based methods [15,16,13,11] have become a very powerful alternative to designing hand-engineered perception and control software, even for basic tasks like collision avoidance. However, a major drawback with such data-driven approaches is that knowledge is usually built from scratch, and often involve complex data acquisition and training procedures.

In this work, we argue that for many robot tasks it is not even possible to obtain real training data. For example, to train an expensive robotic system for collision avoidance using imitation learning we would also need to obtain examples of failure labels. This may often be catastrophic and dangerous (a crashing helicopter). Thus, it has long been a desirable goal to use alternative means such as synthetic simulations to train models that are effective in the real world. Even for tasks where training data can be obtained, the learned policies only apply to the physical system and environment the model was originally trained on, due to the limited variability of datasets. Moreover, in real-world applications we often encounter changes in dynamic conditions, such as weather and illumination, which change the characteristics of the domain. In all of the above scenarios, a good policy cannot be guaranteed if it is trained by using traditional learning techniques. Therefore, there is incentive in establishing techniques to reduce the labeling cost, typically by leveraging labeled data from relevant source domains such as off-the-shelf datasets or synthetic simulations.

Domain adaptation, a method to formally reduce the domain bias, has addressed this problem [14, 1, 7]. However, to date there have been very few attempts to enhance the transferability of learned policies in the context of autonomous robotics. Even fewer have validated them experimentally through real-world experiments. In this paper, we explore these ideas in the context of vision-based autonomous MAV flight [4] in cluttered natural environments, and evaluate how a single policy learned from labeled data from source domain using domain adaptation methods could effectively enable and accelerate learning onto a new target domain.

## 2  Technical Approach

In this section, we describe our proposed approach for learning transferable policies for autonomous MAV flight. Our work is primarily concerned with navigating MAVs that have very low payload capabilities, and operate close to the ground where they cannot avoid dense obstacle fields. We present a system that allows the MAV to autonomously fly at high speeds of up to 1.5 m/s through a cluttered forest environment, using passive monocular vision as its only exteroceptive sensor.

### 2.1  Learning Reactive Policy using Imitation Learning

Visual features extracted from camera input provide a rich set of information that we can use to control the MAV and avoid obstacles. In previous work, we have proposed an imitation learning based technique [16] to directly learn a linear controller of drone's left-right velocity based on visual input. Given a set of human pilot demonstrations in cluttered forest environments and the corresponding images, $\mathcal{D} = \{x_i, y_i\}$, we train a controller to learn a reactive policy $\pi$ that can avoid trees by adapting the MAV's heading as the drone moves forward. Ross et al. [17] showed that over several iterations, the learner is guaranteed to converge to an optimal policy $\pi_n$, based on previous experience, and mimic the behavior atleast as well as the pilot in these situations. However, a major limitation of this approach is that, it can only deal with the minor domain shift induced from sequential prediction tasks, and does not generalize to new environments seamlessly.

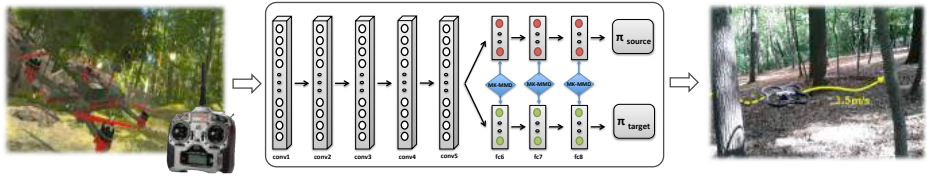

**Fig. 1.** The framework for learning transferable policies from demonstrations in simulated source domain to real target domain using Deep Adaptation Network [12].

### 2.2 Policy Transfer using Deep Domain Adaptation

In this work, we extend the above approach to learn domain-adaptive policies using labeled information from source domain and unlabeled information from target domain. Let the source domain $\mathcal{D}_s = \{(x_i, y_i)\}_{i=1}^{n_s}$ have $n_s$ labeled examples drawn from a probability distribution $p$ and target domain $\mathcal{D}_t = \{(x_j)\}_{j=1}^{n_t}$ have $n_t$ unlabeled examples drawn from a probability distribution $q$. Then, the problem can be formulated as follows: Train a model to learn a set of features $x$ that reduce the cross-domain discrepancy, and a policy $y = \pi_\theta(x)$, where $y$ is the left-right velocity.

Recently, deep convolutional neural network (CNN) based models and features [10] have been proven to be more competitive than the traditional methods on solving complex learning problems. While they have been shown to adapt to novel tasks [6], the main challenge here is that the target domain has no labeled information. Hence, directly adapting CNN to the target domain via fine-tuning is not possible. Thus, we build upon a recently proposed Deep Adaptation Network (DAN) architecture [12], which generalizes deep convolutional neural network to the domain adaptation scenario. The main idea is to enhance domain transferability in the task-specific layers of the deep neural network by explicitly minimizing the domain discrepancy.

In order to achieve this, the hidden representations of all the task-specific layers are embedded to a reproducing kernel Hilbert space where the mean embedding of target domain distributions can be explicitly matched. As mean embedding matching is sensitive to the kernel choices, an optimal multi-kernel selection procedure is devised to further reduce the domain discrepancy. We use a multiple kernel variant of the maximum mean discrepancies (MK-MMD) metric [9] as the measure of domain discrepancy. It is an effective criterion that compares distributions without initially estimating their density functions. The MK-MMD of two distributions $p$ and $q$ is defined as the Reproducing Kernel Hilbert Space (RKHS) distance between mean embeddings of $p$ and $q$:

$$d_k^2(p, q) \triangleq \| \mathrm{E}_p[\phi(x^s)] - \mathrm{E}_q[\phi(x^t)] \|_{\mathcal{H}_k}^2 \tag{1}$$

where $\phi$ is the characteristic kernel associated with the feature map. The most important property here is that $p = q$ iff $d_k^2(p, q) = 0$.

In order to minimize the domain discrepancy in the context of CNNs, we embed a MK-MMD based multi-layer adaptation regularizer (Eq. 1) to the CNN risk function:

$$\min_{\Theta} \frac{1}{n_s} \sum_{i=1}^{n_s} J(\theta(x_i^s), y_i^s) + \lambda \sum_{l=l_1}^{l_2} d_k^2(\mathcal{D}_s^l, \mathcal{D}_t^l) \qquad (2)$$

where $\Theta = \{W^l, b^l\}_{l=1}^{l}$ denotes the set of all CNN parameters, $\lambda > 0$ is the penalty parameter, J is the cross-entropy loss function and $\theta(x_i^s)$ is the conditional probability that CNN assigns $x_i^s$ to label $y_i^s$. $l_1(=6)$ and $l_2(=8)$ are the layer indices between which the regularizer is effective.

The CNN architecture is based on AlexNet [10]. As the domain discrepancy increases in the higher layers [18], we fine-tune the convolutional layers of the CNN (`conv4-conv5`) using source labeled examples and minimize the domain discrepamcy in the fully connected layers (`fc6-fc8`). The deep CNN is trained using a mini-batch supervised gradient descent (SGD) with the above optimization framework (Eq. 2).

## 3  Experiments and Results

In this section, we present experiments to analyze the performance of our proposed method of transferring policies for monocular reactive control of MAVs. All the experiments were conducted in a densely cluttered forest area using the commercially available MAV platforms. We use a distributed processing framework, where the image stream from the front facing camera is streamed to a base station over Wi-Fi at 15 `Hz`. The base station processes these images, and then sends back the desired control commands to the drone.

### 3.1  Methodology

Quantitatively, we evaluate the performance of our system by observing the average distance flown autonomously by the MAV over several runs (at 1.5 m/s), before a crash. Tests were performed in both regions of low and high tree density (approx. 1 tree per $6 \times 6$ $m^2$ and $3 \times 3$ $m^2$, respectively). For each scenario described below, training data with 1 `km` of human-piloted flight was collected in the source domain. Tests were then conducted on approximately 1 `km` of autonomous flights using policies learnt both with and without domain adaptation. As baseline, we compare our results to lower and upper bound: MAV flight using random policy and complete training data, respectively.

### 3.2  Performance Evaluation

**Transfer across systems.** In this experiment we try to answer the question: Can we transfer policies over different physical systems - from one configurations of sensors and dynamics to another? We collect training data using the ARDrone as the source domain, and test on a modified 3DR ArduCopter equipped with a high-dynamic range PointGrey Chameleon camera as the target domain (See Fig. 2a). The sensor system - global shutter vs rolling shutter, image resolution and camera intrinsics are different from that of the ARDrone. Hence, a policy learnt on one system cannot be expected to trivially generalize to the other.

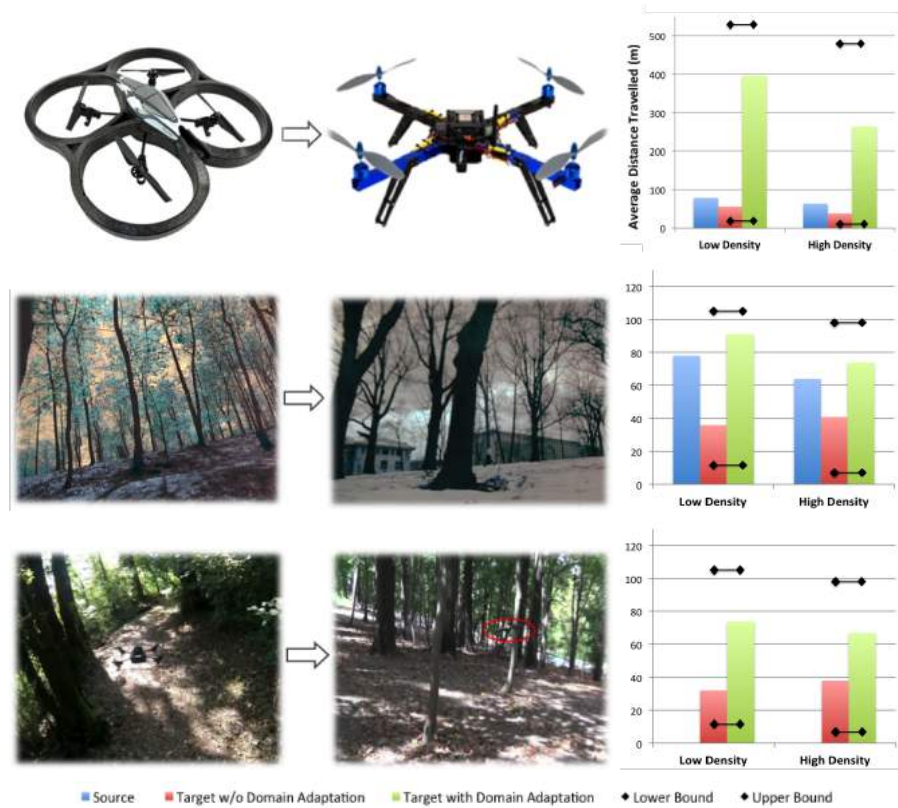

**Fig. 2.** Experiments and Results for (Row-1) Transfer across physical systems from ARDrone to ArduCopter, (Row-2) Transfer across weather conditions from summer to winter and (Row-3) Transfer across environments from Univ. of Zurich to CMU.

**Transfer across weather conditions.** In this experiment we try to answer the question: Can we transfer policies over different weather conditions - from summer to winter? We collect training data during the summer season as our source domain and test during winter season as our target domain (See Fig. 2b). In this case the domain shift is induced by the difference in visual appearance. While the summer environment is cluttered with dense foliage, the winter conditions are often characterized with the presence of snow and absence of foliage.

**Transfer across environments.** In this experiment we try to answer the question: Can we transfer policies over different environments - from one physical location to another? This is equivalent to using an off-the-shelf dataset as the source domain and testing in a seperate target domain. In particular, we use the Univ. of Zurich forest-trail dataset [8] as the source. The dataset provides a large-scale collection of images from a forest environment along with annotations for trail heading (left, center or right). Using these source labels, we train a policy for MAV reactive control and test at the forest environment near CMU as the target domain (See Fig. 2c). Here, the domain shift is induced by the implicit difference in physical location and nature of the task. Note: It is not possible to compare results to the source domain.

### 3.3 Experimental Insights

The main results obtained in this paper is that learning a transferable policy using the proposed approach can boost performance significantly in the target domain, as compared to simply re-using the learnt policy in new domains. Quantitatively, we show this through extensive outdoor flight experiments over a total distance of 6 km in environments of varying clutter density. Even without any training data, in the target domain, the MAV was able to successfully avoid more than 1900 trees, with an accuracy greater than 90%.

We extend our evaluations to qualitatively assess the learned policies during one of the runs from our flight test, as shown in Fig. 3. We show the nature of training data from summer, snapshots of predicted left-right velocity commands in the chronological order of the flight path taken by the MAV. Moreover, we also analyze the the policy learnt without domain adaptation by predicting control commands (offline) using the snapshot images as input. It can be observed that the domain adaptive policy performs better and is able to generalize to the new domain.

Furthermore, we observe that for the first two experiments the performance in the target domain is better than that in the source domain. For transfer across physical systems, this can be attributed to the underlying dynamics of the MAV. The ArduCopter has accurate and stable positioning system that allows it to be more resistive to strong winds, which is a major cause of crash for the less stable ARDrones. Moreover, the target domain has a better sensor suite. The increased resolution probably helped in detecting the thinner trees. For transfers across weather condition, we again observe a boost in performance in the target domain. Empirical analysis of the failure cases reveal that percentage of failures due to branches and leaves diminishes greatly in winter conditions resulting in better overall performance. In comparison to the above two experiments, the performance improves only slightly for transfer across environments. The reason for this is that the source labels are very coarse and were collected for a classification task (left, right or center).

Hence, we learn that to improve the transferability of motion policies over physical systems, it is also important to explicitly address (1) The domain shift induced by discrepancy in dynamics, (2) The expected failure cases in the target

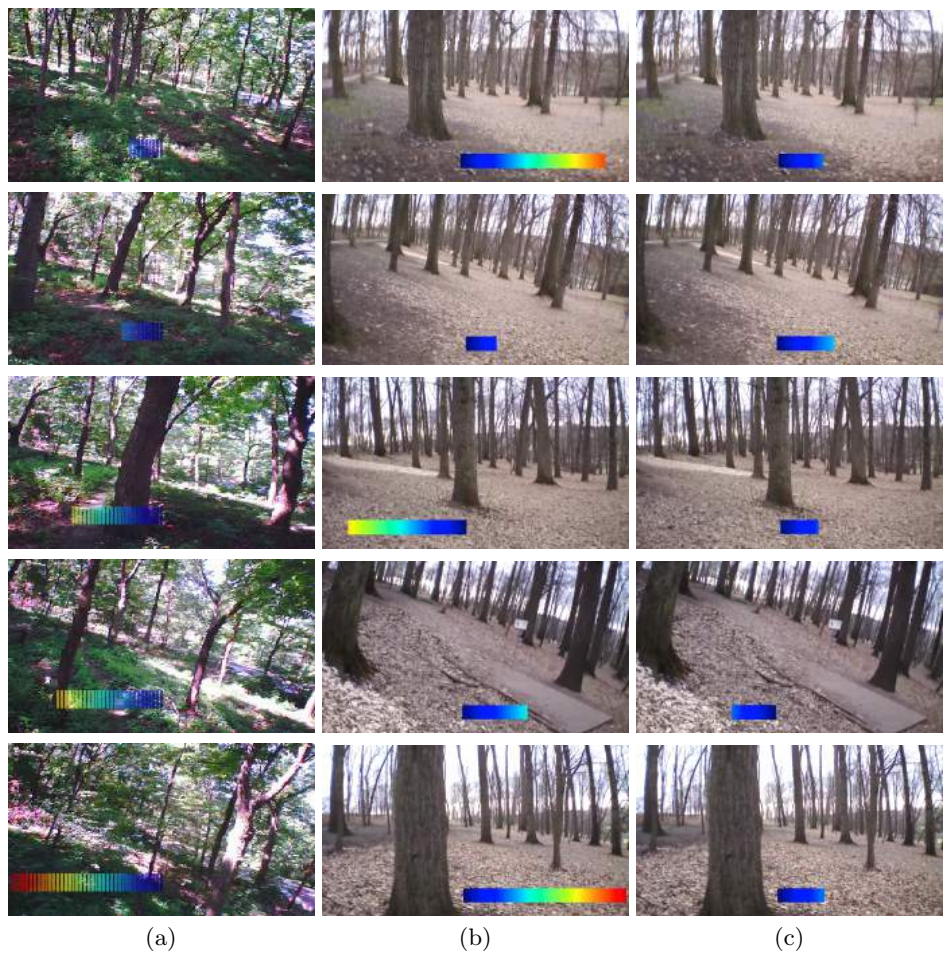

|  (a)  |  (b)  |  (c)  |

**Fig. 3.** Qualitative visualization of an example flight in dense forest. The training data was collected from the same environment during summer season (Col-1) and tested during the winter season (Col-2). The image sequence of MAVs on-board view is chronologically ordered from top to bottom and overlaid with color-coded commands issued by the policy learned using our proposed approach. Additionally, we also compute the commands that would have been generated by the policy without domain adaptation (Col-3), for qualitative comparison.

domain, and (3) The discrepancy induced by not only the domain, but also the task.

## 4  Conclusion

In this paper, we have presented a generic framework for learning transferable motion policies. Our system learns to predict how a human pilot would control a MAV in a source domain, and is able to successfully transfer that behaviour to a new target domain. Quantitatively, we show significant boost in performance over simply re-using policies without any explicit transfer, through extensive real-world experiments. We have demonstrated our approach on an autonomous MAV navigation task using monocular reactive control. However, our treatment and findings apply to any aspect of experimental robotics where a system needs to be trained for end-to-end autonomous behaviour based on sensor data.